\documentclass[12pt]{article}
\usepackage[utf8]{inputenc}
\usepackage[T1]{fontenc}
\usepackage[english]{babel} 

\usepackage{ifpdf,newtxtext,newtxmath} 
\usepackage{array,graphicx,dcolumn,multirow,hevea,abstract,hanging,fancyhdr,float}
\newcommand{\jref}{http://journal.sjdm.org/vol16.1.html}
\newcommand{\jhead}{Role of explanability}
\newcommand{\jdate}{April 2022}
\fancypagestyle{firstpage}{%
 \lhead{\protect\small  \jhead, \jdate}
 \rhead{}
}
\usepackage[labelfont=sc,textfont=sf]{caption}
\usepackage[hyperfootnotes=false,breaklinks=true]{hyperref} 
\usepackage{xcolor}
\usepackage{graphicx}
\usepackage{tikz}
\usepackage{breqn}
\usetikzlibrary{positioning}

\usepackage[hyphenbreaks]{breakurl}

\usepackage{booktabs} 

\newcolumntype{L}[1]{>{\raggedright\arraybackslash }p{#1}} 
\newcolumntype{C}[1]{>{\centering\arraybackslash }p{#1}}
\newcolumntype{R}[1]{>{\raggedleft\arraybackslash }p{#1}}
\newcolumntype{d}[1]{D{.}{.}{#1}} 

\renewcommand\arraystretch{1.2}

\let\tempone\itemize
\let\temptwo\enditemize
\let\tempthree\enumerate
\let\tempfour\endenumerate
\renewenvironment{itemize}{\tempone\setlength{\itemsep}{0pt}}{\temptwo}
\renewenvironment{enumerate}{\tempthree\setlength{\itemsep}{0pt}}{\tempfour}

\title{The Role of Explainability in Assuring Safety of Machine Learning in Healthcare}

\author{
Yan Jia\thanks{Department of Computer Science, University of York, York, UK.}\;\,
\and 
John McDermid \footnotemark[1] 
\and
Tom Lawton \thanks{Bradford Royal Infirmary and Bradford Institute for Health Research, Bradford, UK.} 
\and Ibrahim Habli\footnotemark[1]
}
\date{} 

\begin{document} 
\begin{htmlonly}
\href{\jref}{\jhead}, \jdate, pp.\
\end{htmlonly}

\maketitle
\thispagestyle{firstpage}

\begin{abstract}

Established approaches to assuring safety-critical systems and software are difficult to apply to systems employing ML where there is no clear, pre-defined specification against which to assess validity. This problem is exacerbated by the “opaque” nature of ML where the learnt model is not amenable to human scrutiny. Explainable AI (XAI) methods have been proposed to tackle this issue by producing human-interpretable representations of ML models which can help users to gain confidence and build trust in the ML system. However, little work explicitly investigates the role of explainability for safety assurance in the context of ML development. This paper identifies ways in which XAI methods can contribute to safety assurance of ML-based systems. It then uses a concrete ML-based clinical decision support system, concerning weaning of patients from mechanical ventilation, to demonstrate how XAI methods can be employed to produce evidence to support safety assurance. The results are also represented in a safety argument to show where, and in what way, XAI methods can contribute to a safety case. Overall, we conclude that XAI methods have a valuable role in safety assurance of ML-based systems in healthcare but that they are not sufficient in themselves to assure safety. 

\end{abstract}


\setlength{\baselineskip}{16pt plus.2pt}

\section{Introduction}

In healthcare, machine learning (ML) is used on various problems, e.g. learning optimal treatments, or to detect abnormalities in radiology images, where it has achieved outstanding performance. However, assuring safety for such systems employing ML remains a challenge. In many domains there are well-established approaches and standards for assuring safety-critical systems and software. Assurance means establishing \textit{justified confidence }in the system for its intended use. The assurance principles underlying these standards include validating that the system works as intended and verifying that the system meets explicit \textit{safety} requirements. These assurance principles remain essential for systems employing ML. However, the details of these approaches and standards can be difficult to apply where systems use ML.

First, the established approaches are based, implicitly or explicitly, on the \textit{V life-cycle model} moving from requirements, through design onto implementation then testing. In contrast, the development of ML-based systems follows a very different, much more iterative, life-cycle  with four main  phases: data  management, ML algorithm selection, model learning, and model  verification \& validation, which makes it hard to apply established methods. Some emerging standards and guidance better reflect the ML life-cycle, e.g. the US Federal Drug Administration (FDA) proposed regulatory framework on AI/ML-based Software as a Medical Device (SaMD) \cite{us2019proposed} and Assurance of Machine Learning for Autonomous Systems (AMLAS) \cite{hawkins2021guidance}. 

Second, because of the ``black box'' (opaque) nature of the ML models \cite{watson2019clinical}, it is hard to assess what has been learnt, which exacerbates the challenging of defining concrete requirements for the safety of SaMD in its clinical context. Instead, human performance is often used as a ``gold standard'' and the current practice is often to (seek to) achieve performance that is better than humans. This makes validation difficult as human performance is variable both from individual-to-individual and over time for a single individual. Also, performance will vary from patient-to-patient, e.g., with comorbidities, and clinicians might not agree on the best treatment strategy.

To overcome such problems, the ML community is actively studying ``explainablity'', which is intended to ``peek inside the black box'' and to illuminate the underlying workings of the ML models. Explainability is often equated with producing explainable artificial intelligence (XAI) methods, which seek to provide human interpretable representations of ML models \cite{challen2019artificial}. Although there is considerable variation in the definition of terms such as explainability, interpretability and transparency, in this paper we adopt the view from the FDA AI/ML-enabled Medical Devices Transparency Workshop \cite{fdadefination} that explainability is one component of transparency. Transparency is a much broader concept in their definition and we see interetability as a necessary facet of explainability, as suggested by Gilpin et al \cite{gilpin2018explaining}.

In this paper we consider the role of explainability in assuring the safety of ML models in healthcare. Our focus is on development activities and deployment decisions for ML-based systems, but we briefly consider the potential role of explainability in operations.

The primary contributions of this paper are as follows. First, we developed a new conceptual model, a spider diagram, which gives a heuristic view of safety and shows how safety, as a cross-cutting concern, relates to other aspects of an ML-based system, including the role of explainability. Then we show how to assure safety by meeting relevant regulatory requirements, especially from the FDA. In particular, we show how XAI methods can be integrated into the different phases of model development and what types of XAI methods best provide safety evidence to meet the regulatory requirements.  Finally, we present a concrete case study to illustrate the use of XAI methods in supporting a safety case for ML systems. 

The rest of the paper is structured as follows. Section \ref{background} identifies relevant related work. Section \ref{Expl in LCM} outlines the potential role of XAI methods in the ML life-cycle. These possibilities are then investigated in Section \ref{example} using an example of weaning patients from mechanical ventilation. This is followed by a discussion and conclusions in Sections \ref{discussion} and \ref{conclusions}, respectively. 

\section{Background \& Related Work}
\label{background}

This section discusses established approaches to assurance of safety-critical systems and identifies their limitations when dealing with systems employing ML. This is followed by an introduction to the concepts of explainability and an overview of the different types of XAI methods.

\subsection{Established Assurance Approaches and the Challenges of ML}
\label{established}

We use the term \textit{assurance} to mean \textit{confidence that the system behaviour is as intended in the environment of use}, where \textit{as intended} includes being safe. In this context, we are interested in assurance of patient safety when ML-based systems are used in a healthcare context. 

Most approaches to assurance emphasise verification and validation, although the definitions of the terms can vary. The  International Medical Devices Regulator Forum (IMDRF) define the terms as follows: 

\begin{itemize}
    \item Verification -- confirmation through provision of objective evidence that specified requirements have been fulfilled \cite{imdrf2017software};
    \item Validation -- confirmation through provision of objective evidence that the requirements for a specific intended use or application have been fulfilled \cite{imdrf2017software}.
\end{itemize}

To interpret these definitions we can say that validation is concerned with \textit{building the right system}, including defining requirements that meet our intent and that verification is concerned with \textit{building the system right} by demonstrating that the system meets these requirements. 
Verification and validation (V\&V) need to encompass identified safety requirements, which are often derived to control the risk associated with \textit{hazards}, i.e. undesirable situations that pose risk to life. Typically, risk is a combination of the likelihood and the severity of the harm arising from the hazard, although the detailed computations vary from domain to domain. Where risks are deemed too high, derived safety requirements (DSRs) are identified to reduce the likelihood of hazard occurrence, e.g. by controlling hazard causes, or to mitigate the consequences of the hazard should it arise. In healthcare, such ideas underpin some of the relevant standards, e.g. \cite{DCB} produced by NHS Digital in the UK.

We have previously investigated how to adapt traditional safety engineering processes to healthcare systems which employ ML. We have shown that, in some cases, it is possible to adapt classical safety methods to identify hazards and then to establish DSRs on the ML elements of systems \cite{jia2021safety}. Clinical judgement is also needed to produce such DSRs. Where requirements are not stated explicitly, XAI methods can help by providing explanations that enable direct validation of the ML model as a whole, e.g. showing that predictions are based on valid clinical factors and are consistent with clinical knowledge.

In many domains, including healthcare, it is accepted good practice for the safety work to culminate in the production of a safety or assurance case, see \cite{DCB}. In general, a safety case is ``an argument, supported by evidence, that a systems is safe to be deployed in its context of use''. It is common to express the argument graphically, e.g., using the Goal Structuring Notation (GSN) \cite{GSNCommu77:online}, as a means of making the argument clear and open to review. The evidence underpinning the safety case includes the results of hazard and risk analysis, as well as the outputs from V\&V activities and in this paper we will show that XAI methods can also provide such evidence. 


There are a number of initiatives concerned with the assurance of ML in safety-critical systems both in healthcare and more generally. For example, AMLAS defines a process for assurance of the safety of ML-based systems to reflect the ML development life-cycle, which identifies both evidence artefacts and argument patterns (standard forms of argument that can be instantiated for a particular system) in GSN. AMLAS also considers issues of the robustness of ML-based systems, e.g. response to unexpected inputs. The FDA also proposed a total life-cycle regulatory approach for ML-based SaMD \cite{us2019proposed}. However, these approaches are evolving in that they provide good high-level guidance and objectives, but how to meet such objectives is not sufficiently detailed. The work we present here is intended to be complementary to, and build on, these approaches and shows how XAI methods can provide evidence to meet these objectives, and thus contributes to improving their maturity.


In addition, it is always desirable to consider assurance ``through life'', as proposed by the FDA \cite{us2019proposed}, not just as an activity undertaken prior to deployment. This includes getting feedback from operations to check whether or not the assumptions made in pre-deployment assurance activities are sound. This is even more important for ML-based systems than it is for ``conventional'' systems because of the opacity of ML models. 


\subsection{Explainable AI Methods}
\label{XAI methods}

ML includes a range of different methods such as decision trees, support vector machines and neural networks (NNs). The study of XAI methods seeks to provide insight into how and why ML models make such predictions. Work on explainable AI includes formalising definitions of explainability, development of XAI methods themselves and establishing evaluation methods. In this section we provide a brief overview of XAI methods. For a more complete view of XAI, please refer to some well-cited surveys, e.g. \cite{arrieta2020explainable}, \cite{gilpin2018explaining}. There are many different ways to categorise XAI methods, e.g. local or global based on the scope of the explanation or model agnostic or specific based on whether the XAI methods can work for any class of ML models or only work for a specific class of models. Here we adopt the taxonomy of XAI in \cite{arrieta2020explainable} and present XAI methods in two different classes based on the explanation generating mechanism, as shown in Table \ref{tab:XAI-method}. 

Some ML models are perceived as intrinsically interpretable to the user, so we refer to these as \textit{interpretable models}. This includes linear/logistic regression, decision trees, K-nearest neighbours, decision rules, Bayesian models, general additive models (GAMs), etc. Note that, although often these models are viewed as intrinsically interpretable, when the number of input features are beyond human ability to grasp or when the input features are heterogeneous which is not uncommon in healthcare, it can be difficult for humans to interpret the model and care needs to be taken \cite{mood2010logistic}. 

When it comes to explaining more complex or opaque ML models, e.g. support vector machines, tree ensembles, and NNs, which are not intrinsically interpretable, a \textit{post-hoc} explanation can be used to provide insights without knowing the mechanisms by which the model works. 
We present four main \textit{post-hoc} explanation classes as follows along with some popular techniques as illustrations.

\begin{itemize}
    \item \textbf{Explanation by approximation} aims to use surrogate models, e.g. linear models, decision trees or decision rules to approximate the underlying complex or opaque model. These can include local or global surrogates depending on whether they are approximating a single prediction or the whole model. For example, \textbf{LIME} (Local Interpretable Model-Agnostic Explanations)  \cite{ribeiro2016should} focuses on training a local surrogate to provide explanations for an individual prediction, which is based on the assumption that it is possible to fit a surrogate model around a single input sample that mimics the local behaviour of the complex ML model. Like LIME, \textbf{Anchors} \cite{ribeiro2018anchors} deploy a perturbation based strategy to generate local explanations for predictions in a local region resulting in if-then rules. In contrast, \textbf{model extraction} proposed in \cite{bastani2017interpretability} trains a global surrogate to approximate a complex model. The three methods mentioned above are model-agnostic so they would also work for NNs.
    \item \textbf{Explanation by example} explains the ML model by selecting particular instances from the dataset or by creating new instances. It comprises counterfactual examples, adversarial examples, influential instances and prototypical examples. \textbf{Counterfactual examples} \cite{wachter2017counterfactual} can be thought of as ``what is not, but could have been'' and are intended to produce a sparse human-interpretable example by changing some input features to achieve a different output, i.e. the user's desired output (what could have been). \textbf{Adversarial examples} \cite{szegedy2013intriguing} are typically generated by adding small, intentional perturbations to the input features to cause an ML model to make an incorrect prediction. Adversarial examples are intended to deceive the ML model instead of interpreting the model. Therefore, the changes in the inputs are often imperceptible for a human observer. \textbf{Influential instances} are intended to identify which input instances have a strong effect on the trained model. They can be identified by measuring the impact of a training point on a particular prediction or on the model overall. \textbf{Prototypical examples} can summarise and represent a complex underlying  data distribution, which then can be used to provide a global understanding of the model by examining prototypes along with their model predictions or a local explanation for a specific instance by identifying the most similar training instance according to the trained model. This is different to influential instances because the training example might be influential but not representative. 
    
   \item \textbf{Feature relevance explanation} techniques rank or score the input features based on their influence, relevance or importance on the model prediction where higher scores mean that the corresponding features are more important for the model. Such scores are often obtained by perturbation or gradient-based methods. For example, \textbf{SHAP} (SHapley Additive exPlanations) \cite{lundberg2017unified} is one of the perturbation methods based on Shapley values, which are used to explain a model prediction by treating input features as the players in a cooperative game and the model prediction as the gain resulting from the game. It includes KernelSHAP, a model agnostic weighted linear regression approximation of the exact Shapley value inspired by LIME, and TreeSHAP, a model-specific efficient estimation approach for tree-based models. The work on SHAP has wider significance as it has defined a new class of additive feature importance measures, unifying several existing XAI methods \cite{lundberg2017unified}.
    \item \textbf{Visual explanation} techniques aim to facilitate model understanding by using visualisation, e.g. showing how features interact with the predicted output or other features. Such methods can involve using sensitivity analysis (SA) or partial dependence to inspect the relationship between the uncertainty in the predicted output and its input features, e.g. \cite{CORTEZ20131} presented several visualisations for the SA results and \cite{goldstein2015peeking} introduced Individual Conditional Expectation (ICE) to show how the prediction of a particular instance changes along with the input features.
\end{itemize}

Due to the popularity of deep learning (DL), a different classification scheme for XAI in DL is proposed by Gilpin et al \cite{gilpin2018explaining}, and it is often treated as a subfield of XAI \textit{per se}.
\begin{itemize}
    \item \textbf{Explanations of deep network processing}. This can be achieved by producing a ``saliency map'' which is a rendering of weights for the input features that highlight the salient features for the prediction. A ``saliency map'' can be produced by perturbation-based methods, e.g. LIME, or by calculating the gradient of the output with respect to the input, identifying which parts of the input have a significant influence on the classification \cite{simonyan2013deep}. Due to some limitations of directly using gradients, e.g. saturation, there are also a number of other methods proposed, for example, Integrated Gradients \cite{sundararajan2017axiomatic}, LRP \cite{bach2015pixel}, and DeepLIFT \cite{shrikumar2017learning}.    
    \item \textbf{Explanations of deep network representations}. This type of explanation aims to inspect what the model learnt. Feature visualisation \cite{olah2017feature} is helpful to understand how an NN builds up its understanding of input images throughout the network by maximising activation for the unit of interest, e.g. a specific neuron, or a specific layer, or a convolution channel. Further, there are concept-based methods, e.g. TCAV \cite{kim2018interpretability}, attempting to detect concepts that are human-interpretable but embedded within the latent space learnt by the network. 
    \item \textbf{Explanation-producing systems}. For example, attention-based networks learn a function by providing weights of the input or internal features of a NN in order to force the model to attend to the important regions with respect to the target task. Although the attention-mechanism could render an attention map to provide some insights or intuitive feeling of the model, it is important to note that the interpretation of attention as explanation is currently the subject of debate \cite{jain2019attention}\cite{wiegreffe2019attention}.
\end{itemize}

Some further details on XAI methods we have used in our case study are included in Section \ref{example}.

\begin{table}[!ht]
\renewcommand{\arraystretch}{1.3}
\caption{Categorisation of XAI Methods with Examples}
\label{tab:XAI-method}
\centering
\resizebox{0.7\textwidth}{!}{%
\begin{tabular}{|ll|l|l|l|}
\hline
\multicolumn{2}{|l|}{Type of explanation} & Scope & \begin{tabular}[c]{@{}l@{}}Model \\ Specific/\\ Agnostic\end{tabular} & Examples of XAI methods \\ \hline
\multicolumn{2}{|l|}{Interpretable Models} & Global & Specific & \begin{tabular}[c]{@{}l@{}}A model by itself interpretable,\\ e.g. linear/logistic regression,\\ decision tree, GAM, etc.\end{tabular} \\ \hline
\multicolumn{1}{|l|}{\multirow{14}{*}{\begin{tabular}[c]{@{}l@{}}Post-hoc\\ explanation\end{tabular}}} & \multirow{3}{*}{\begin{tabular}[c]{@{}l@{}}Explanation by \\ Approximation\end{tabular}} & Local & Agnostic & LIME \\ \cline{3-5} 
\multicolumn{1}{|l|}{} &  & Local & Agnostic & Anchors \\ \cline{3-5} 
\multicolumn{1}{|l|}{} &  & Global & Agnostic & Model extraction \\ \cline{2-5} 
\multicolumn{1}{|l|}{} & \multirow{4}{*}{\begin{tabular}[c]{@{}l@{}}Explanation by\\ example\end{tabular}} & Local & Agnostic & Counterfactual examples \\ \cline{3-5} 
\multicolumn{1}{|l|}{} &  & Local & Agnostic & Adversarial examples \\ \cline{3-5} 
\multicolumn{1}{|l|}{} &  & Local & Agnostic & Influential instances \\ \cline{3-5} 
\multicolumn{1}{|l|}{} &  & Local & Agnostic & Prototypical examples \\ \cline{2-5} 
\multicolumn{1}{|l|}{} & \multirow{5}{*}{\begin{tabular}[c]{@{}l@{}}Feature \\ relevance\\ Explanation\end{tabular}} & Local & Agnostic & KernelSHAP \\ \cline{3-5} 
\multicolumn{1}{|l|}{} &  & Local & Specific & TreeSHAP \\ \cline{3-5} 
\multicolumn{1}{|l|}{} &  & Local & Specific & LRP \\ \cline{3-5} 
\multicolumn{1}{|l|}{} &  & Local & Specific & Integrated Gradient \\ \cline{3-5} 
\multicolumn{1}{|l|}{} &  & Local & Specific & DeepLIFT \\ \cline{2-5} 
\multicolumn{1}{|l|}{} & \multirow{2}{*}{\begin{tabular}[c]{@{}l@{}}Visual\\ Explanation\end{tabular}} & Global & Agnostic & Partial dependence plot \\ \cline{3-5} 
\multicolumn{1}{|l|}{} &  & Local & Agnostic & ICE \\ \hline
\end{tabular}
}
\end{table}

\section{The Role of Explainability in Safety Assurance}
\label{Expl in LCM}

\begin{figure}[!ht]
    \centering
    \includegraphics[scale=0.4]{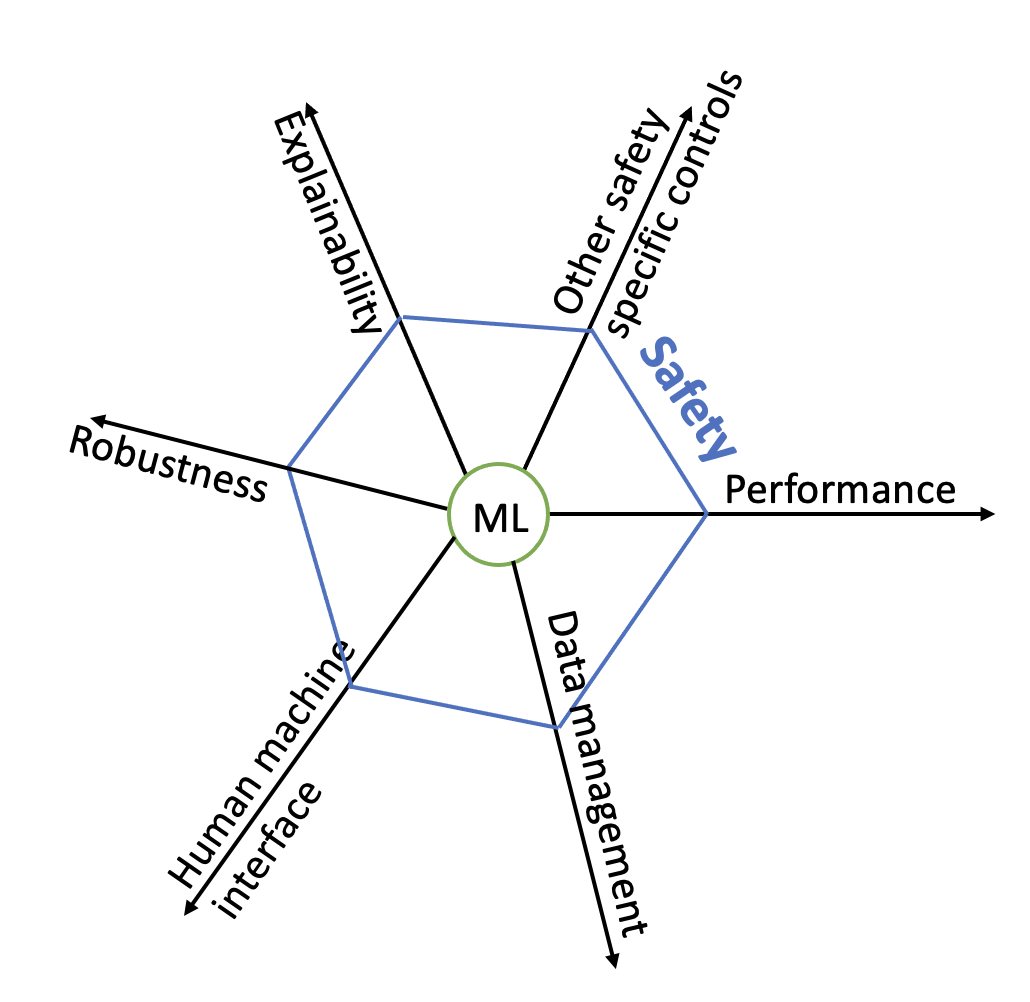}
    \caption{Spider diagram illustrating the role of explainability and safey}
    \label{spider_diagram}
\end{figure}
Assurance of safety is a multi-faceted, multi-dimensional concern. We start by presenting our new conceptual model as a spider diagram to show the role of explainability in safety assurance generally, and how this relates to regulatory requirements. Then, we exemplify how explainability can contribute to safety assurance in the context of the ML life-cycle.

\subsection{Explainability and Safety}

Fig. \ref{spider_diagram}, which we refer to as the \textit{spider diagram}, shows the different aspects of a ML-based system, e.g. performance and explainability, that relate to safety in its context of use. This illustrates that safety is a cross-cutting concern, rather than being a separate dimension. Fig. \ref{spider_diagram} gives an impression of safety, where safety is related to six dimensions of ML-based systems and the area within the hexagon can be viewed as a heuristic evaluation of safety. Note that the dimensions in Fig. \ref{spider_diagram} are not exhaustive; we have chosen them as they provide a good engineering perspective.

\textit{Performance} of the ML can be assessed using accuracy or other metrics, which are well known. \textit{Data management} for ML is crucial as the quality of data has a significant impact on the model learnt, e.g. if the data is biased, then the model might also learn discriminatory behaviour or even amplify it.  \textit{Human-machine interface} is also very important for safety as certain types of interface can be prone to human operator errors. Here we think of \textit{Robustness} as the model prediction being stable, particularly in the presence of small variations in the input features. \textit{Explainability} can be achieved using intrinsically interpretable models or \textit{post-hoc} XAI methods. However, explanations of intrinsically interpretable models are more accurate than \textit{post-hoc} XAI methods, in the sense that they have high fidelity to the task model, and this has led to some authors suggesting that only such models should be used for safety-critical tasks \cite{rudin2019stop}. Finally, \textit{Other safety specific controls} can be thought of as means to satisfy DSRs at the system level, which are not covered by the above dimensions. This is task dependent, e.g. in an online healthcare triage system, it is important to have a safety net where specific words should trigger an emergency response rather than continuing to ask questions. As shown in Fig. \ref{spider_diagram}, performance also matters for safety but if an intrinsically interpretable model can achieve similar performance to a ``black box'' model, then the interpretable one should be preferred.



To draw the spider diagram for a particular ML-based system requires metrics for the different dimensions. Performance is readily quantified, but it is less obvious how to measure or quantify the other dimensions; we return to this issue in Section \ref{discussion}. In theory, if we could identify all the relevant dimensions and quantify the area in the spider diagram this would give a good basis for safety assurance. In practice, regulators are defining regulatory requirements with the intent that if these requirements are satisfied, the system can be approved as safe enough to be marketed.

In the next subsection, we focus on explanability and further investigate how it can contribute to safety in the context of the ML life-cycle, and show how the evidence generated by XAI methods can contribute to meeting the relevant FDA regulatory requirements.

\subsection{Explainability in the ML life-cycle}
\begin{figure}[!ht]
\centering
\includegraphics[width=0.7\textwidth, clip=true, trim=0cm 2.6cm 3cm 1cm]{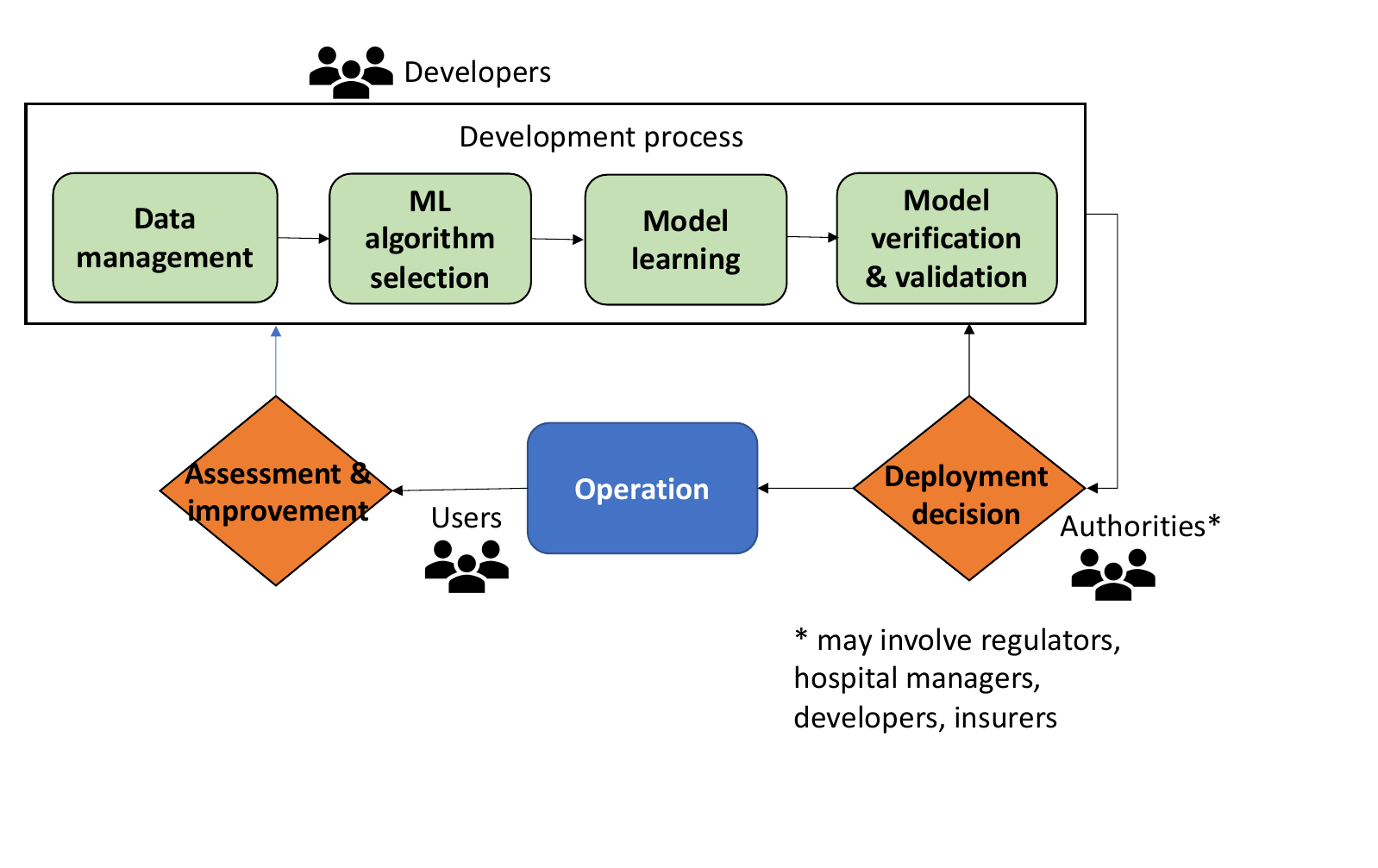}
\caption{Process for development and use of an ML System} \label{fig2}
\end{figure}

The development process for ML typically includes data management, ML algorithm selection, model learning and model V\&V \cite{hawkins2021guidance}, as shown in Fig. \ref{fig2}. Fig. \ref{fig2} makes explicit the need for a deployment decision prior to operation (which may be supported by a safety case). It also shows the stakeholders who might be interested in the explanations in the different phases. Our focus here is on the development activities, but we briefly consider the potential role of explainability in operation, see Section \ref{Operation}; for a discussion of the wider role of explainability including incident and accident investigation see \cite{mcdermid2021philtransa}.

The rest of this section discusses the role of explainability for each stage of the process shown in Fig. \ref{fig2}. 

\subsubsection{Data Management}

The first phase of the ML development process is data management, and this aligns directly with the spider diagram. Most XAI methods are not applicable to this stage but \textit{prototypes} are relevant; they can help to understand the datasets especially when the datasets are large and complex, although \textit{prototypes} can also be used to approximate the learnt model \cite{tan2020tree}.

The Royal Society's Policy Briefing on XAI emphasises that data quality and provenance is part of the explainability pipeline, specifically saying that \textit{``Understanding the quality and provenance of the data used in AI systems is therefore an important part of ensuring that a system is explainable''} \cite{policybriefing}. This includes showing that the data comes from appropriate sources for the problem addressed. A widely accepted, harmonised framework for assessment of Electronic Health Record (EHR) data quality highlights \textit{conformance}, \textit{completeness} and \textit{accuracy} \cite{kahn2016harmonized}; we prefer \textit{accuracy} to the original term plausibility because plausibility means that the values are in the possible range but accurate means that the data is not only possible but correct. These criteria would be applicable to any ML systems developed using EHR data. Further, we also identify data \textit{relevance} and \textit{balance} as being particularly important to ML model development \cite{hawkins2021guidance}. As real world data may contain biases, errors, or be incomplete, explaining how these five criteria are met can be at least as important as explaining the ML model itself.

The evidence to ensure data quality is essentially technical, for example data \textit{conformance} would include showing that data observes defined formats, e.g. correct units for weight \cite{kahn2016harmonized}. However, demonstrating \textit{data relevance} and \textit{data balance} would include a judgement that the training data contained clinically relevant factors and are balanced for the problem being addressed. We acknowledge that often it is not possible to choose data that gives both feature balance and class balance. Instead, it might be useful to explain that some important features are reasonably balanced, e.g. gender, if the model is intended to be used for both male and female patients. Class balance has long been an active research area in the ML community. In the case of skewed dataset, \textit{prototypes} can be generated to understand the data distribution and can be used to train a model. There is evidence showing that using \textit{prototypes} can help with class balance. For example, Gurumoorthy et al \cite{gurumoorthy2019efficient} have demonstrated that using good \textit{prototypes} to train a model can give better performance than using the whole dataset or randomly sampled subsets to balance the classes. 

It should be noted that data management is both crucial and labour intensive. Indeed, it may consume more effort than the rest of the ML life-cycle. Judgement of the extent to which the data meets these five criteria would be used to assess an AI/ML system in the \textit{data management} dimension in the spider diagram.


\subsubsection{ML Algorithm Selection}
\label{ModSel}

The second phase in the development process is ML algorithm selection (also referred to as model selection, here we use the term ML algorithm selection to avoid the confusion with model selection in the training phase where the ML algorithm is the same but hyperparameters of the model are tuned to be different). It is important to understand what kind of problem is being addressed and what kind of ML methods are suitable for the problem at hand, e.g. classification, regression, or finding an optimal policy. Another important aspect to consider at this stage is the \textit{explainability} and \textit{performance} dimensions of the ML model as shown in the spider diagram. In Section \ref{XAI methods} we identified that some ML models are intrinsically interpretable whereas others need to be supplemented with \textit{post-hoc} XAI methods. Guidelines on ML algorithm selection, balancing model performance against explainability, have been proposed \cite{markus2020role}. 

When it comes to ML algorithm selection, safety requirements are often implicitly transformed into explainability and performance requirements. Note that sometimes people make statements such as ``use of deep NNs is not safe''. When they make this kind of statement, they are implicitly making the judgement that deep NNs are opaque, i.e. not interpretable. The spider diagram helps to show that this is over-simplistic. There is no binary choice ``opaque/interpretable'', rather \textit{explainability} and \textit{performance} are two of the dimensions; both matter for safety in an AI/ML application. This is why we argue that safety requirements are partially, but not wholly, transformed into explainability requirements. It would be ideal to have an interpretable model which can achieve performance as high as black box models. When this is not the case, a trade-off between explainability and performance would be necessary \cite{markus2020role} and \textit{post-hoc} explanations should be considered either in later phases of development or in operation to produce effective explanation. The rationale for the ML algorithm choice, including the performance-explainability trade-offs, needs to be documented in the safety case.



\subsubsection{Model Learning}
\label{MLVaV}

The third phase in the development process is model learning. The essential aim of this stage is to train a ``good'' model, and \textit{performance} and \textit{robustness} are especially important at this stage. For model learning, hyperparameter selection, loss function definition and class balance need to be considered in order to meet safety requirements. In addition, XAI methods can have a role in improving both \textit{performance} and \textit{robustness}, e.g. through model debugging and adversarial training. 


\textbf{Automated robustness improvement} -- \textit{adversarial examples} are often added to training data to improve model robustness in object classification tasks. This is referred to as adversarial training or robustness training \cite{ayers2020parot} \cite{huang2015learning}. This is becoming widespread in domains such as autonomous vehicles, for example in improving performance at reading road signs under adverse conditions \cite{eykholt2018robust}, but we believe it has wider applicability, e.g. for image classification in radiology. There are many techniques for generating adversarial examples, e.g. by minimising the distance between the adversarial example and the input instance, which is similar to counterfactual examples (see Section \ref{ModelVandV} for more discussion). Popular distance metrics include: $L_{0}$, $L_{1}$, $L_{2}$ and $L_{\infty}$, all of which are $L_{p}$ norms \cite{carlini2017towards}. $L_{0}$ counts the number of features that have changed between the two instances, $L_{1}$ measures the sum of the magnitudes of the change, $L_{2}$ measures the Euclidean distance between the two instances, and $L_{\infty}$ measures the maximum change among all of the features.
Use of \textit{adversarial examples} to improve model robustness is becoming widespread in domains such as autonomous vehicles, for example in improving performance at reading road signs under adverse conditions, but we believe it has wider applicability, e.g. for image classification in radiology.

\textbf{``Model debugging''} -- \textit{influential instances} are useful as they can help to understand model behaviour, specifically they can help to debug domain mismatch or fix problematic training instances where the label or input features might be incorrect \cite{koh2017understanding}. We can investigate influential instances for a specific prediction or for the model overall. For example, when the test instances are being misclassified by the model, we can identify the most responsible training instances for such instances. This enables exploration of the causes of the problem, e.g. errors in the training instances, or domain mismatch, i.e. poorly represented subgroups in the dataset, which can therefore be unduly influential. Domain mismatch is not uncommon in healthcare as the population from the intended use hospital can be very different from the population used for developing the model. Further, we can also investigate the most influential training instances for the model overall, e.g. we can assess the influence of removing a certain training instance on the model by measuring the change in loss of this training instance before and after it is removed; the bigger the change in the loss the more influential the instance. Intuitively if the model has to ``try hard'' to accommodate the instance, then that instance is highly influential. In reality, labels in the training dataset can be noisy and it is unrealistic to expect human experts to manually review all of the data. In this case, it is useful to investigate the most influential instances for the model overall to direct the human experts' attention to the instances that actually matter.

Two approaches for identifying \textit{influential instances} are often used -- deletion diagnostics and influence functions. Deletion diagnostics is not practical for big training datasets as it needs to remove a single training instance every time to observe the effect of this instance until the effect of all of the training data has been observed. However, influence functions can be used to approximate the effect without deleting the training instance (see Section \ref{CasestudyLearning} for details).

In addition to the uses of the two XAI methods described above, current research is also exploring how to guide the learning process to enable the models to produce the desired form of explanations, for example by including explainable regularisers in the loss function. In \cite{ross2017right}, the authors penalise the gradient of a NN to force it to focus on regions which contain important information for the task. Others are exploring how to penalise the neighbourhood fidelity \cite{plumb2019regularizing} in order to improve the quality of the explanation. This is an important research direction, but it relies on understanding what constitutes a good explanation and what metrics enable the ``goodness'' of the explanation to be evaluated, e.g. neighbourhood fidelity or stability \cite{plumb2019regularizing}. We expect such methods to be significant for safety assurance, once the understanding of what is a good explainable regulariser is more mature, which depends on progress in the domain of explainability itself.


\subsubsection{Model Verification and Validation}
\label{ModelVandV}

The final phase in the development process is model V\&V. We believe that explainability has a particular role in validation, but could also have a role in verification if there are specific explainability requirements to verify. However, such explainability requirements need to be defined in a specific situation, therefore our focus here is on validation. 
We derived three distinct objectives, reconciling approaches proposed by the FDA \cite{us2019proposed} and the IMDRF \cite{imdrf2017software}, which reflect key criteria for use of ML models in healthcare, although we note that explanations cannot guarantee that all these criteria are met \cite{markus2020role}. 

First is \textit{performance}, which can be measured using standard ML practices, e.g. evaluation of the proportion of false positives and false negatives, or the AUC-ROC. This is necessary but not sufficient to assure safety of ML.

The second objective is \textit{analytical} or \textit{technical validation}, showing that the software for the ML models is correctly constructed, and that it is accurate and reliable. Further, the ML model implementations produce repeatable results, giving the same predictions from the same inputs. This objective can be met by employing established safety-critical software development practices including formal specifications, traceability from specification to implementation, use of test coverage criteria and static code analysis methods \cite{McDermid2020SafetyOA}. We do not see a role for XAI methods for this aspect of validation.


Third is \textit{clinical validation} which measures the ability of the system to generate a clinically meaningful output for its intended use in its operational environment. Here we define two specific sub-objectives where we believe XAI methods have a role in supporting clinical validation: 
\begin{itemize}
    \item \textbf{Clinical association} -- demonstrate that the association between the system output and the targeted clinical condition in the intended population is supported by evidence;
    \item \textbf{Robustness} -- demonstrate the ability to distinguish the different classes of intended condition or recommended treatment without over-reliance on a specific input feature.
\end{itemize}

These explicitly relate to the \textit{explainability} and \textit{robustness} dimensions in the spider diagram.

Feature relevance explanations can help to demonstrate clinical association by showing that the output predictions are based on clinically meaningful and relevant factors of the input. This involves ranking input features based on their importance score or contribution score and making the rankings visible to clinicians so that they can exercise clinical judgement. One might argue that rule-based explanation can also help in this case, for example, using Anchors to generate relevant rules for clinicians to judge whether a valid clinical association has been learnt. However, for the following reasons, we believe that feature relevance explanations should be preferred in this context. First, clinicians consistently stated that knowing the feature importance driving the model outcome is crucial. This allows them to compare model decisions to their clinical judgement, especially in case of a discrepancy \cite{tonekaboni2019clinicians}. Second, ML is most valuable when it is used in complex clinical tasks where there are no agreed set of rules. In this case, it is easier to agree on the important and unimportant factors than on rules. For example, there is evidence \cite{kovalerchuk2001consistent} showing that even if an extracted rule provides 100\% accuracy in the test dataset, an individual clinician still might not have confidence in it, let alone achieving consensus amongst the clinicians. Third, using feature importance can facilitate the regulatory approval process, and this avoids hampering the adoption of ML until universal rules are found and agreed by experts. However, we acknowledge that rule based methods are useful for knowledge discovery. Feature relevance explanation also has its limitations in explaining image-based datasets where the highlighted regions in a saliency map might not correspond well to high level concepts that are meaningful to humans. In this case, concept-based approaches, e.g. TCAV \cite{kim2018interpretability}, that can quantify the degree to which a user-defined concept is important to a prediction, would be more appealing. In healthcare, such concepts would be pre-defined by clinicians.

Example-based explanation, especially \textit{counterfactual examples}, can help to assess model robustness. Counterfactuals are generated by minimising the distance from the original input whilst producing a different prediction. As we mentioned in Section \ref{MLVaV}, the distance metrics for adversarial examples are still valid for counterfactuals, although among them $L_{1}$ is the most widely explored in the literature and $L_{\infty}$ is rarely used, which is unsurprising as $L_{1}$ can enforce sparsity in the generated example. As pointed out in \cite{mothilal2020explaining}, sparsity is one of the desirable properties of a good counterfactual. Intuitively, a counterfactual will be more implementable, and easier to understand, if fewer features have to be changed. Additional desirable properties for good counterfactuals are proximity (as close as possible to the original instance), plausibility (it is possible for the features to take that value) and diversity (multiple ways of achieving the desired prediction). On-going research seeks novel loss functions to incorporate these properties to generate the counterfactuals. When using counterfactuals to assess model robustness, proximity is the most important property, in other words if the counterfactual methods don't satisfy this property, they should not be used to assess robustness. Intuitively, the greater the distance from an initial input to a counterfactual, the more robust the ML model is, i.e. the model is ``harder to fool''. Therefore, distance metrics can be used to define a robustness score for the ML model, which links to the robustness dimension of the spider diagram. For example, \cite{sharma2019certifai} used the $L_{1}$ distance metric to define the score. However, we suggest that the $L_{2}$ distance metric might be most appropriate to use based on the findings in \cite{carlini2017towards} which indicates that achieving robustness against $L_{2}$ also achieves robustness against other distance metrics.



The use of XAI methods in support of ML model V\&V will contribute evidence to the safety case, complementing other activities including performance assessment and safety-critical software engineering. It should be noted that explanations should be re-generated when the ML models are updated so that they reflect the state of the models.

\subsubsection{Operation}
\label{Operation}

As discussed in Section \ref{established}, assurance should be considered to be a ``through life'' activity. This would include, for example, a clinician seeking assurance about a particular prediction, especially if acting on it can have a profound impact on patient safety. XAI methods can play a role here. Local feature relevance explanation may be helpful but counterfactuals also have a role, e.g. helping a clinician to decide whether or not a proposed change in treatment is likely to bring about the desired effect for a particular patient. Further, prototypes might also be able to help clinicians to make informed decisions for specific patients. However, current research, e.g. \cite{tan2020tree}, often presents prototypes to laymen rather than domain experts to assess whether this will help them to make decisions, so more work needs to be done to see whether clinicians can benefit from prototype explanations. The role and significance of explainability in operation is examined in more detail in \cite{mcdermid2021philtransa}. 

\section{Case Study}
\label{example}

This section presents a concrete healthcare case study to illustrate the role of XAI methods, introduced in Section \ref{Expl in LCM}. The case study doesn't cover data management, but see our previous work for an illustration of the rationale for data inclusion for this case study \cite{jia2021}. The case study focuses on use of mechanical ventilation in Intensive Care Units (ICUs). Provision of mechanical ventilation is complex and consumes a significant proportion of ICU resources \cite{ambrosino2010difficult}. Invasive mechanical ventilation is used when patients cannot breathe unaided, and requires the insertion of a tube into the trachea of the patient. The term intubation is used for insertion of tube and extubation for removal of the tube. 
It is of critical importance to determine the right time to wean the patient from mechanical ventilation. Both early and late weaning are problematic. Early extubation can lead to the need for re-intubation, which may become urgent. Late extubation exposes a patient to discomfort and continued risk of complications such as pneumonia from prolonged intubation. 

The case study is particularly concerned with predicting patient readiness for extubation so as to avoid the negative side effects of mis-timed extubation using the features shown in Fig. \ref{whole_patient}. Put simply, the safety requirement is ``prediction of readiness for extubation is timely''. The  case study is based on the MIMIC-III dataset \cite{johnson2016mimic} and used a convolutional NN (CNN) to predict readiness for extubation in the next hour. 


\subsection{ML Algorithm Selection}
\label{modelsele}

ML Algorithm selection is strongly influenced by performance, as previously indicated. There are a range of performance metrics, e.g. false positives, which in this case study would mean indicating that a patient is ready for extubation when it was actually premature. Here we use the AUC-ROC performance measure. The ROC curve plots the true positives against the false positives at various threshold settings. AUC-ROC represents the degree to which the model is capable of distinguishing between classes. For a ``random'' model the AUC-ROC would be 0.5 and for a ``perfect'' model it would be 1. 

For the case study, the performance of a number of ML models, including CNNs, were evaluated on the same dataset to support this phase, see Table \ref{fig:modelsele}. CNNs have the best performance and more importantly, achieve better performance than decision trees and logistic regression which, as noted above, are often viewed as intrinsically interpretable. As mentioned in section \ref{ModSel} there is a trade-off between performance and explainability. If performance over-rides the need for explainablility, then CNN should be chosen. Whilst if intrinsic interpretability is more important, then logistic regression should be chosen. In this case study, CNNs have been chosen, and \textit{post-hoc} XAI methods are used to explain the model, see the rest of the section for details. 



\begin{table}[!ht]
\renewcommand{\arraystretch}{1.3}
\centering
\caption{Performance of ML models}
\label{fig:modelsele}
\begin{tabular}{|l|l|}
\hline
\textbf{Model}               & \textbf{AUC-ROC (95\%CI)} \\ \hline
Convolutional Neural Network & 0.923 $\pm$ 0.010             \\ \hline
Artificial Neural Network    & 0.784 $\pm$ 0.031             \\ \hline
Logistic Regression          & 0.827 $\pm$ 0.000             \\ \hline
Random Forest                & 0.748 $\pm$ 0.040             \\ \hline
Decision Tree                & 0.808 $\pm$ 0.008             \\ \hline
Support Vector Machine       & 0.826 $\pm$ 0.000             \\ \hline
\end{tabular}
\end{table}

\subsection{Model Learning}
\label{CasestudyLearning}
As we indicated in Section \ref{MLVaV}, two XAI methods can be helpful at this stage: adversarial examples and influential instances. Because adversarial examples are difficult to generate for tabular data, here we focus on the use of influential instances for ``model debugging''. This shows how they provide assurance about the appropriateness of the ML model learning process, in the context of the safety requirement.

When preparing the dataset for the case study, one issue that came up was whether or not to include the extubation failure patients. Here extubation failure is defined as the need for re-intubation within 48 hours. The causes of extubation failure are complex and unclear, but some of the literature suggests that premature extubation could cause extubation failure \cite{ambrosino2010difficult}.
Therefore, including extubation failure patients in the training dataset might not be optimal, as it might negatively influence the prediction. To explore this issue further, we trained two CNN models to predict the readiness for extubation in the next hour in order to observe the effect of extubation failure patients. In the first model, we excluded all of the extubation failure patients in the training dataset. In the second model, we included all of the extubation failure patents in the training dataset. The accuracy of the second model is slightly changed by comparison with the first model. We randomly picked one of the test instances that was ``interesting'' in that the two models produced different predictions. For this instance, the first model predicted the patient should continue to be intubated, which is also the true label. However, the second model predicted that the patient was ready for extubation in the next hour. We used influence functions to identify the influential training instances for this test instance.

The key idea behind influence functions is to up-weight the loss of a training instance by an infinitestimally small step $\epsilon$, which results in new model parameters, $ \hat{\theta}_{\epsilon,z}$ = $argmin(1-\epsilon)\frac{1}{n}\sum_{i=1}^{n}L(z_i, \theta) + \epsilon L(z,\theta)$,
where $\theta$ is the model parameter vector and $\hat{\theta}_{\epsilon,z}$ is the model parameter after upweighting $z$ by $\epsilon$. $L$ is the loss function used for training the model. 
The influence of upweighting $z$ on the parameters $\hat{\theta}$ given by Cook and Weisberg \cite{cook1982residuals} is as follows:

\begin{equation}
    I_{up,params}(z) = \frac{d\hat{\theta}_{\epsilon,z}}{d\epsilon}|_{\epsilon=0} = -H_{\hat{\theta}}^{-1}\nabla_{\theta}L(z,\hat{\theta}) 
\end{equation}

Where $H_{\hat{\theta}}$ is the Hessian matrix and $\nabla_{\theta}L(z,\hat{\theta})$ is the loss gradient with respect to the parameters $\hat{\theta}$ for the training instance $z$. Next, we can apply the chain rule to calculate the influence of upweighting instance $z$ on the loss of a test instance $z_{test}$:
\begin{dmath}
    I_{up,loss}(z, z_{test}) = \frac{dL(z_{test}, \hat{\theta}_{\epsilon,z})}{d\epsilon}|_{\epsilon=0} =  \nabla_{\theta}L(z_{test},\hat{\theta})^{T}\frac{d\hat{\theta}_{\epsilon,z}}{d\epsilon}|_{\epsilon=0} =
    -\nabla_{\theta}L(z_{test},\hat{\theta})^{T} H_{\hat{\theta}}^{-1}\nabla_{\theta}L(z,\hat{\theta}) 
\end{dmath}

\begin{figure}[!ht]
	\begin{tikzpicture}
  \node(img) {\includegraphics[scale=.3]{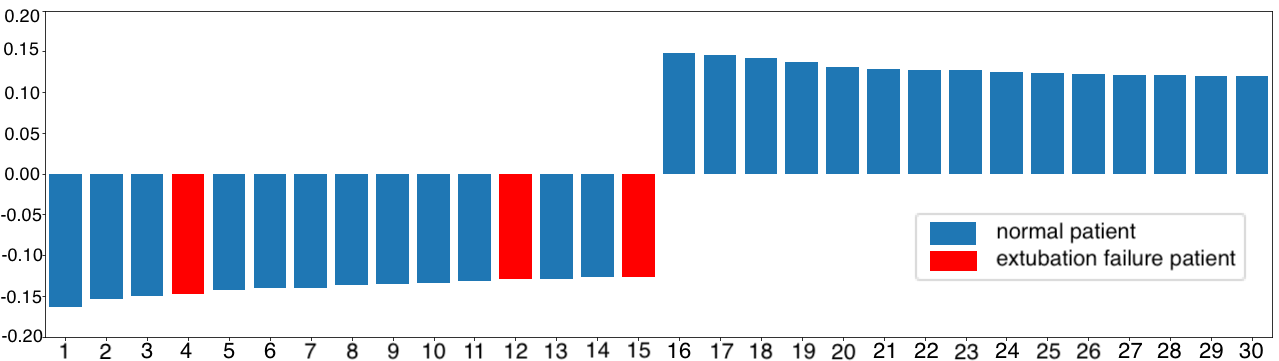}};
  \node[below=of img, node distance=0cm, yshift=1.1cm,font=\color{black}] {\scriptsize \textit{Training instances}};
  \node[left=of img, node distance=0cm, rotate=90, anchor=center,yshift=-0.95cm,font=\color{black}] {\scriptsize $-I_{up,loss}$};
 \end{tikzpicture}
	\caption{Top 30 most influential training instances}
	\label{influential_top30}
\end{figure}

\begin{figure}[!ht]
	\begin{tikzpicture}
  \node(img2) {\includegraphics[scale=.3]{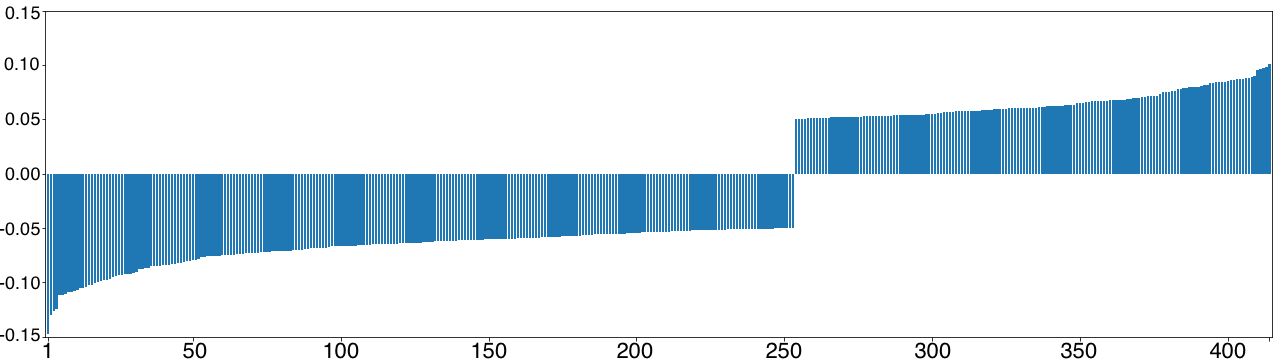}};
  \node[below=of img2, node distance=0cm, yshift=1.1cm,font=\color{black}] {\scriptsize \textit{Training instances}};
  \node[left=of img2, node distance=0cm, rotate=90, anchor=center,yshift=-0.95cm,font=\color{black}] {\scriptsize $-I_{up,loss}$};
 \end{tikzpicture}
	\caption{Distribution of influential instances}
	\label{influential_414}
\end{figure}

In this work, we use the influence functions algorithm developed by Koh and Liang \cite{koh2017understanding} to calculate $-I_{up,loss}(z_i, z_{test})$ for each training instance $z_i$ for this test instance. Fig. \ref{influential_top30} shows the top 15 helpful training instances (most positive $-I_{up,loss}(z_i, z_{test})$) and the top 15 harmful training instances (most negative $-I_{up,loss}(z_i, z_{test})$) for this test instance.
 From the figure, it shows there are three instances of patients who had extubation failure among the harmful training instances, which indicates that including the extubation failure patients made the predictions for the test instance worse. Fig. \ref{influential_414} shows some of the most influential data points (magnitude of $-I_{up,loss}(z_i, z_{test})$ is large) from the extubation failure patients and that more of them have a negative influence than a positive influence. This suggests that 
 the inclusion of extubation failure could make the prediction ready to extubate when it is not the case. Thus, we decided to exclude the extubation failure patients from the training dataset and the first CNN model was taken forward to the V\&V stage. In a more general situation when prior knowledge is not available, i.e. we don't know what subset of the data could be problematic, we can still choose a test instance where the prediction is wrong and identify the influence of the training instances on this prediction. Then, further investigation could be done to understand what input features strongly impact the influence score, e.g. by perturbation \cite{koh2017understanding} or by using decision trees \cite{molnar2020interpretable}.


\subsection{Model Verification and Validation}
\label{CasestudyVandV}

In this section, we focus on clinical validation, as set out in Section \ref{ModelVandV}, and illustrate the use of XAI methods for demonstrating clinical association and robustness. We do not consider analytical validation here. 
	
\subsubsection{Feature relevance explanations}
\label{FeatImp}


Here we illustrate the role of feature relevance in satisfying the clinical association safety assurance objective. This is done using DeepLIFT \cite{shrikumar2017learning} which is a model-specific XAI method for deep NNs. It compares the activation of each neuron to its ``reference activation'' and attributes to each input feature an importance score based on the difference. The ``reference activation'' is obtained through some user-defined reference input and in this case, the reference sample is the minimum values of all of the input features obtained from the data set. We chose this method for two main reasons. First, it deals effectively with discontinuities in the gradient of the CNN model as it uses a difference from reference approach. Second, it avoids the problem of model saturation where using gradients would just assign zero to the features \cite{shrikumar2017learning}.

\begin{figure}[!htbp]
	\centering
		\includegraphics[scale=.5]{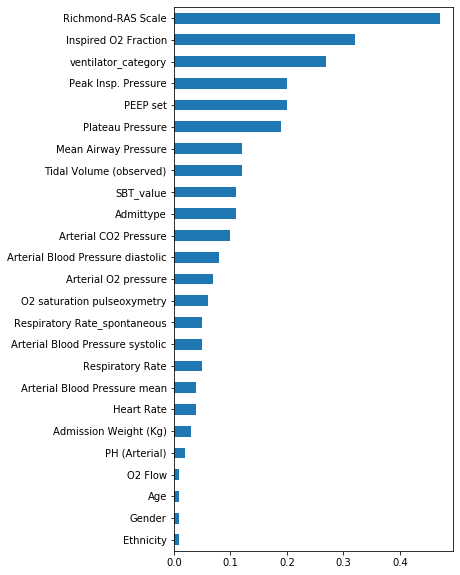}
	\caption{Feature Importance for the CNN Model}
	\label{whole_patient}
\end{figure}

An overview of the results of using DeepLIFT is shown in Fig. \ref{whole_patient}; these values are averaged over the whole dataset, so this can be viewed as global feature importance. The feature ranking correlates well with clinical expectations, helping to give confidence in the model. 
Those features that score near zero in Fig. \ref{whole_patient}, e.g. ethnicity, gender and age, have little influence on the weaning decision, which is as expected. The top five features also align with clinical evidence. Patients who are undergoing invasive mechanical ventilation are often sedated to maintain physiological stability and to control pain levels. Sedation is reflected in the Richardson Agitation Scale (RAS) with negative values representing sedation and 0 meaning that they are alert and calm, thus  more  likely to be suitable for extubation. This is consistent with the first entry in the weaning checklist used in \cite{walsh2004evaluation} that patients are ``cooperative and pain free''. The second most important feature is ``Inspired O2 fraction'' which is the third checklist entry in \cite{walsh2004evaluation}. The third most important feature is ``ventilator category'', which is the mode used for ventilation and is under direct clinician control; some modes are unsuitable for spontaneous breathing so cannot easily support weaning. The fourth and fifth most important features, peak inspiratory pressure and positive end-expiratory pressure (PEEP) set are airway pressures representing how hard the ventilator is having to work; PEEP is also the third entry in the weaning checklist in \cite{walsh2004evaluation}.

Here we have demonstrated valid clinical association through clinical evidence (relevant literature support) and expert opinion (consultation with clinicians). Overall, the benefit of the feature importance results is that they enable clinical judgement to be applied despite the opacity of the CNN model which contributes to safety assurance. 

\subsubsection{Counterfactual explanations}
\label{CountFact}

\begin{table}[!htbp]
\renewcommand{\arraystretch}{1}
\centering
\caption{Counterfactual examples for a given instance }
\label{tab:my-table}
\resizebox{0.7\textwidth}{!}{
\begin{tabular}{lllll}
\hline
\multirow{2}{*}{Features}    & \multirow{2}{*}{Original instance} & \multicolumn{3}{l}{Counterfactual Examples} \\
                           &           & 1 & 2 & 3  \\
\hline
Admit Type                 & Emergency & --- & ---  & --- \\
Ethnicity                  & White     & ---  & ---  & --- \\
Gender                     & Female    & ---  & ---  & --- \\
Age                        & 78.2      & ---  & ---  & --- \\
Admission Weight           & 86.5      & ---  & ---  & ---  \\
Heart Rate                 & 119       & ---  & 110  & ---  \\
Respiratory Rate           & 24        & 26   & ---  & ---  \\
SpO2                       & 98        & ---   &---  & 96   \\
Inspired O2 Fraction       & 100\%     & ---  & 40\%  & ---   \\
PEEP set                   & 10        & 5    & 5   & 5  \\
Mean Airway Pressure       & 14        & ---  & 10  & ---    \\
Tidal Volume (observed)    & 541       & ---  & ---   & 560   \\
PH (Arterial)              & 7.46      & ---  & ---  & ---   \\
Respiratory Rate(Spont)    & 0         & ---  & 24   & --- \\
Richmond-RAS Scale         & -1        & ---  & 0  &  --- \\
Peak Insp. Pressure        & 21        & ---  & ---  & ---  \\
O2 Flow                    & 5         & ---  & ---  & ---    \\
Plateau Pressure           & 19        & ---  & ---  & ---  \\
Arterial O2 pressure       & 124       & 108 & 118  & ---   \\
Arterial CO2 Pressure      & 33        & ---  & ---  & ---   \\
Blood Pressure (systolic)  & 101       & ---  & ---  & ---   \\
Blood Pressure (diastolic) & 65        & ---  & --- & ---   \\
Blood Pressure (mean)      & 76        & ---  & --- & ---    \\
Spontaneous breathing trials & \begin{tabular}[c]{@{}l@{}} No result \end{tabular} &
\begin{tabular}[c]{@{}l@{}} Successfully \\ Completed \end{tabular} &
\begin{tabular}[c]{@{}l@{}} Successfully \\ Completed \end{tabular} &
\begin{tabular}[c]{@{}l@{}} Successfully \\ Completed \end{tabular} \\ 
Ventilator Mode & 
\begin{tabular}[c]{@{}l@{}} CMV/ASSIST/ \\ AutoFlow \end{tabular} &
\begin{tabular}[c]{@{}l@{}} PCV+  \end{tabular} &
\begin{tabular}[c]{@{}l@{}} SIMV/PSV  \end{tabular} &
\begin{tabular}[c]{@{}l@{}} SIMV/PSV  \end{tabular} \\ \hline
Predicted outcome          & 0.93      & 0.44  & 0.17  & 0.36  \\
\hline
\end{tabular}
}
\end{table}

One of the concerns in model V\&V is robustness and here we show how to use counterfactuals to demonstrate robustness. Diverse Counterfactual Examples (DiCE) \cite{mothilal2020explaining} is used to generate the counterfactual examples. The reason we use DiCE is that it is one of the few methods to satisfy all of the four properties of a good counterfactual, i.e. sparsity, proximity, plausibility and diversity, introduced in Section \ref{ModelVandV}. Table \ref{tab:my-table} shows a set of counterfactual examples for a particular patient identifying which features need to change in order to ``flip'' the prediction from continued intubation to extubation using DiCE. The left hand column shows the 25 features used by the model and the prediction of the ML model is included in the bottom row. The original instance is shown first, with the three rightmost columns showing counterfactual examples. Certain features cannot be varied, e.g. age and gender, in order to satisfy the plausibility criterion; the dashes in the rightmost three columns indicate no change from the original input. The change in prediction is shown in the bottom row where a value $>$0.5 indicates that mechanical ventilation should continue.


In this case, as shown in Table \ref{tab:my-table}, the minimum number of features that have to change to ``flip'' the prediction is five, showing robustness for this instance. However, one instance is not sufficient to show ML model robustness. More of the input instances in the dataset need to be investigated in order to generate a robustness score as defined in \cite{sharma2019certifai}.

\subsection{Operational use of the ML Model}
\label{opML}

The operation of ML models is often uncertain. Thus there is merit in extending the notion of assurance to operation, providing support to a clinician to give confidence to act on the particular model prediction. One way of approaching this is to use local explanations. For example, we can generate the feature importance for a specific patient, similarly to Fig. \ref{whole_patient}.

However, clinicians might want to find out when the patient would be ready to extubate. This brings us back to counterfactuals. The counterfactual examples shown in Table \ref{tab:my-table} could potentially help the clinician to identify actions to take so that the patient becomes ready to extubate. The model does not directly recommend a course of action; the counterfactual examples act to draw clinicians' attention to pertinent information so that they can formulate a plan from their own knowledge and experience. Note our model has not been used in operation yet, so we have just illustrated the possibilities.

\subsection{Safety arguments}
\label{MLSA}

As explained in Section \ref{established}, it is common practice to present the arguments and evidence that provide assurance that a system is acceptably safe to deploy in a safety case. In this case study, the safety argument is presented using GSN. Before we describe the safety argument we have developed, we briefly introduce the notation.
\begin{figure}[!ht]                     
  \centering                      
  \includegraphics[scale=1]{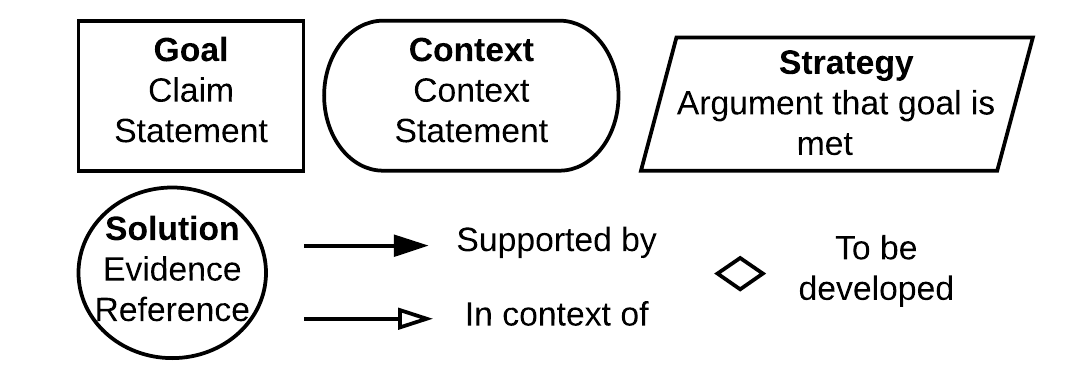}      
   \caption{Goal Structuring Notation}   
   \label{gsnlegend}
\end{figure} 
\begin{figure*}[!ht]
	\centering
		\includegraphics[scale=.7, clip=true, trim=0cm 0.3cm 0cm 0cm]{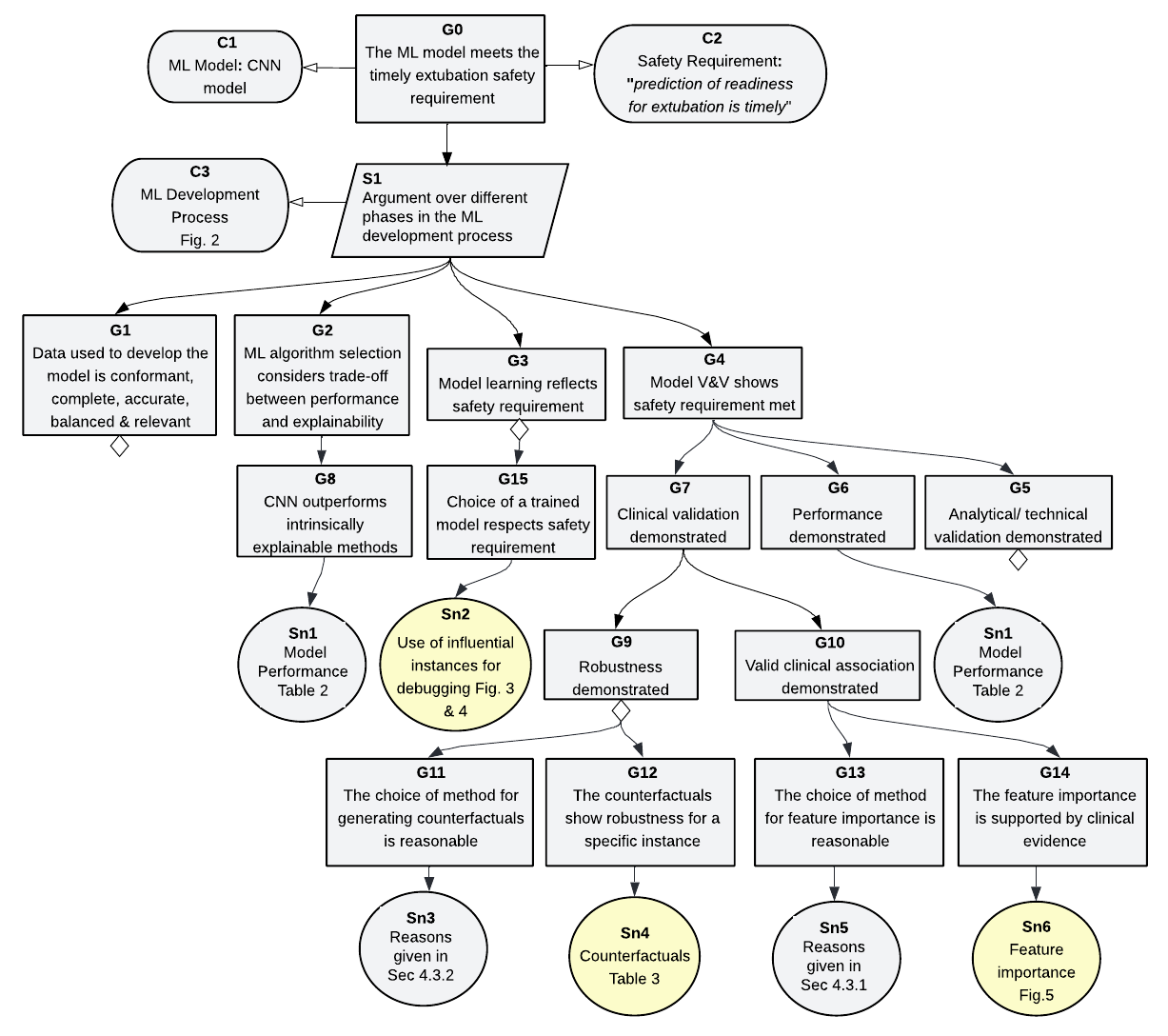}
	\caption{Partial Safety Argument for Weaning ML Model emphasising Explainability}
	\label{SafetyArgument}
\end{figure*}

A legend showing the key elements of GSN is presented in Fig. \ref{gsnlegend}; a detailed description of the notation can be found in \cite{GSNCommu77:online}. The \textit{goals} -- claims that we wish to make and support -- are shown as rectangles and they can be decomposed into sub-goals, thus forming a tree. \textit{Goals} are understood in a \textit{context}, e.g. the operating environment for the system or the safety requirements. Where the decomposition of goals is not obvious this is explained through a \textit{strategy}, represented as a rhombus. The leaf-level goals are supported by \textit{solutions}, represented as circles; the solutions provide references to evidence that supports the argument. Incomplete parts of the argument are shown with a diamond, meaning that part of the argument is to be developed.  

Fig. \ref{SafetyArgument} presents a partial safety argument for the weaning case study, highlighting the role of explainability, e.g. solutions Sn2, Sn4 and Sn6 reflect XAI methods. The top goal (G0), which states that the ML model meets its safety requirement, is set out in the context of the definition of the ML model and the associated safety requirement -- that ``prediction of readiness for extubation is timely''. 

The top-level argument strategy is decomposition across the stages of the ML development process. As the paper does not consider data management and analytical/technical validation in detail, the corresponding goal (G1) and goal (G5) are left undeveloped. 

G2: ML algorithm selection considers trade-off between performance and explainability -- this is supported by the analysis in Section \ref{modelsele} (Sn1) which shows that the CNN outperforms other available ML methods, and suitable \textit{post-hoc} XAI methods are available. 

G3: Model learning reflects safety requirement -- this is \textit{partially} supported by G15  which 
in turn is supported by the use of influential instances (Sn2) which show the rationale for excluding extubation failure patients in the trained model. Note that other evidence is needed (so G3 is shown as needing development), e.g. to show appropriateness of parameter selection for model training.

G4: Model V\&V shows safety requirements met -- this is broken down into G5: analytical/technical validation, G6: performance demonstrated and G7: clinical validation which is decomposed into two sub-goals covering the V\&V criteria introduced in Section \ref{ModelVandV}. 

G6: Performance demonstrated -- this is directly supported by the AUC-ROC in Table \ref{fig:modelsele} which shows the superiority of the CNN performance to others.

G9: Robustness demonstrated, which is decomposed into two subgoals, G11 concerning the reason for the choice of the specific XAI method and G12 concerning the results of using the method. It is important to document the rationale for choosing the method, which will benefit from a more systematic evaluation metric for XAI methods, see Section \ref{discussion} for details. G9 needs further development as G12 only shows robustness for a single instance. More input instances need to be investigated to understand the model robustness. 

G10: Valid clinical association demonstrated, which is decomposed into two subgoals, G13 concerning the reason for the choice of the specific XAI method and G14 concerning the results of using the method. G14 is further supported by clinical evidence in the literature and expert review by clinicians. Meeting G10, demonstrating valid clinical association, does not entail trust. Trust is a more complicated topic \cite{markus2020role} and is outside the scope of this paper. 

The evidence presented above should not be taken as sufficient to justify deployment of the CNN model described here in a clinical context. For example, clinical trials will often be needed to obtain further clinical safety evidence and the necessary regulatory approval. However, the explainability argument and the supporting evidence presented in this section is a valuable part of the overall safety case.

\section{Discussion}
\label{discussion}

Safety assurance of ML models in healthcare is an active area of research. Although explainability is often said to help in safety assurance of ML, few studies so far have explored the possibilities systematically and identified precisely how explainability can help safety assurance. This paper seeks to fill this gap. We first presented the spider diagram in Fig. \ref{spider_diagram} which conceptualises how explainability relates to safety. Through the analysis in Section \ref{Expl in LCM} and the case study in Section \ref{example} we illustrated how explainability can help in safety assurance in the context of the ML life-cycle, as summarised in Table \ref{tab:discussion} along with the interested stakeholders. Although using explainability can help safety assurance, there are also associated challenges.

\begin{table}[!ht]
\renewcommand{\arraystretch}{1.3}
\centering
\caption{Role of XAI Methods in different phases }
\label{tab:discussion}
\resizebox{0.7\textwidth}{!}{%
\begin{tabular}{|l|l|l|l|}
\hline
\textbf{Phases} & \textbf{Activity} & \textbf{XAI methods} & \textbf{Stakeholders} \\ \hline
\multirow{4}{*}{Development} & Data Management & Prototypes & \begin{tabular}[c]{@{}l@{}}ML developers\\ Regulators\\ Hospital managers\end{tabular} \\ \cline{2-4} 
 &  \begin{tabular}[c]{@{}l@{}}ML Algorithm \\ Selection\end{tabular}   & \begin{tabular}[c]{@{}l@{}}Trade-off performance \& \\ explainability\end{tabular} & ML developers \\ \cline{2-4} 
 & Model Learning & \begin{tabular}[c]{@{}l@{}}Adversarial examples\\ Influential instances\end{tabular} & ML developers \\ \cline{2-4} 
 & Model V \& V & \begin{tabular}[c]{@{}l@{}}Global feature importance\\ Counterfactual explanations\end{tabular} & \begin{tabular}[c]{@{}l@{}}ML developers\\ Regulators\\ Hospital managers\\ Insurers\end{tabular} \\ \hline
Operation & Decision Support & \begin{tabular}[c]{@{}l@{}}Local feature importance\\ Counterfactual explanations\end{tabular} & \begin{tabular}[c]{@{}l@{}}Expert users:\\ clinicians\\ Decision recipients:\\ patients\end{tabular} \\ \hline
\end{tabular}%
}
\end{table}

First is the difficulty of evaluating XAI methods. A considerable number of evaluation metrics have been suggested for assessing the quality of XAI methods. For explanation by approximation, fidelity is often proposed as the evaluation metric. For example, in \cite{ribeiro2016should} \cite{plumb2019regularizing}, fidelity is used to measure how accurately the approximate model matches the task model locally. For feature relevance explanation, their ideal properties, e.g. implementation invariance (the feature importance are always identical for two functionally equivalent networks) and sensitivity (if a feature  changes and a prediction changes, then this feature should not have zero attribution), have been defined using axiomatic evaluation methods \cite{sundararajan2017axiomatic}. For explanation by examples, humans are often involved in assessing whether the explanation is useful or not. For example, in \cite{tan2020tree} prototypes were presented to users and then human accuracy, i.e.the proportion of the human predictions that correctly match the model's prediction, was measured. 
Among these proposed evaluation metrics, some are subjective and some are objective. One of the useful taxonomies was proposed by Doshi-Velez and Kim \cite{doshi2017towards} for evaluating XAI: application-grounded, human-grounded, and functionality-grounded. However, there is still a lack of agreed formal evaluation metrics enabling a more systematic evaluation of methods. For example, \cite{adebayo2018sanity} found Gradients \& GradCAM passed their sanity checks based on their model parameter randomization test and the data randomization test, while in \cite{KAKOGEORGIOU2021102520} they found GradCAM is one of the most interpretable and reliable XAI methods but gradient didn't stand out based on their evaluation metrics. This highlights the importance of ongoing research developing systematic evaluation metrics which will allow a formal and fair comparison of available XAI methods. This should help to guide the selection of an appropriate XAI method in a specific situation as many different XAI methods can provide similar explanations, e.g. feature relevance.

Second is the limitation of explainability itself. Even if the evaluation metrics are improved, there are some intrinsic limitations of the current XAI methods. As pointed out by Rudin \cite{rudin2019stop} \textit{post-hoc} explanations ``must be wrong'' as if they were completely faithful to the task model, then we only need the explanation model. Further, as illustrated in \cite{ghassemi2021false}, current XAI methods only provide descriptive accounts of the task model rather than normative evaluation to justify the model behaviour. Whilst this is valid, it is unrealistic to expect XAI methods to ``close the loop'' by themselves; instead this is the role of clinical judgement, as discussed above. Put more positively, explanation ``serves as the unacknowledged bridge'' between the task model and normative evaluation \cite{ghassemi2021false}.  What these examples make clear is that, although the use of XAI methods can contribute to safety assurance, it is not enough to assure safety by itself, as recognised by others \cite{markus2020role} \cite{ghassemi2021false}.

Next, we will identify some relevant complementary methods that also contribute to safety assurance. First, it is important to adapt established methods from safety-critical software engineering for AI/ML-based SaMD. One such method is static analysis, analysing the code without executing it, to look for ``bugs'' (see \cite{McDermid2020SafetyOA} for an illustration). It is also standard practice to measure test coverage of the software, e.g. ensuring that all branches in the code have been executed at least once, when undertaking V\&V. The obvious analogy for NNs is neuron coverage, although there is some debate about whether or not this is an appropriate criterion \cite{li2019structural}. Nonetheless, coverage is significant when considering safety, as assurance is clearly undermined if there are significant parts of the ML model for which we have no test evidence. Consequently, it seems likely that understanding of what are appropriate coverage criteria will improve as experience of using AI/ML-based SaMD increases. Second, there are assurance methods that address the specific challenges of V\&V for AI/ML-based software. It is possible to apply formal methods (mathematical techniques of verification) to ML models including for assessing properties such as robustness. A recent survey of such approaches can be found in \cite{urban2021review}; it covers static analysis as well as methods such as Satisfiability Module Theories (SMT) and identifies some tools that are capable of scaling to very large models, e.g. NNs with millions of neurons.

Safety is a cross-cutting concern which interacts with technology, ethics, trust, etc. Earlier we introduced the spider diagram in Fig. \ref{spider_diagram} as a heuristic representation of the engineering aspects of safety; we now briefly consider how to make it more concrete, i.e. how the area inside the hexagon \textit{might} be estimated. There are \textit{candidate} quantitative measures for some of the dimensions, e.g. AUC-ROC (see Table \ref{fig:modelsele}) for performance, and distance metrics to define robustness scores \cite{sharma2019certifai}. There is a conceptual basis for some of the other dimensions, e.g. fidelity and interpretability for explainability \cite{markus2020role}, and conformance with the five criteria for data management which also links to the safety argument. However, as we mentioned earlier more work is needed to develop systematic evaluation metrics for XAI methods. It is hard to measure the other safety specific controls dimension but the structured representation of controls in \cite{jia2021safety} might give a starting point. There is ongoing work on utility of explanations and what form of explanations are preferred by clinicians \cite{lahav2018interpretable}, but more empirical work is needed. At this point, the spider diagram remains a heuristic model, but we believe it is possible to make it a more ``formal'' tool for evaluating (comparing) alternative ways of developing SaMD, from a safety assurance perspective.



\section{Conclusions}
\label{conclusions}
To our knowledge, this is the first systematic attempt to explore the role of explainability in assuring safety of ML, with a focus on pre-deployment decision-making. We believe this will be of interest to regulators, as it illustrates how to use XAI methods to provide evidence to support relevant safety objectives, e.g. for clinical association, articulated by the FDA and IMDRF. 

We first presented a spider diagram to the general relationship between explainability and safety. Then we extrapolated the safety objectives at the different phases of the ML development process and illustrated how XAI methods can help at each phase. Finally, we used a concrete healthcare case study to demonstrate how XAI methods can help to meet these safety objectives, particularly in model learning and model V\&V. Specifically, we have shown the value of influential instances for model debugging during model learning, which is of particular interest to ML developers. Further, we have  shown the value of feature relevance and counterfactuals in model V\&V, which is of particular interest to ML developer, regulators and others involved in deployment decisions, see Table. \ref{tab:discussion}. The case study also shows how the use of these XAI methods feeds into a safety case, e.g. as required by healthcare standards \cite{DCB}. 

We suggest future work should place more effort on developing and applying systematic evaluation metrics for XAI methods, which in turn will guide others in selecting appropriate XAI methods. Further, there is a need for more empirical studies to evaluate how XAI methods can best assist the end-users. This might usefully be combined with a deeper exploration of trust.


The code for applying various XAI methods is available at: https://github.com/ Yanjiayork/mechanical\_ventilator.

\section*{Acknowledgement}
This work is funded by Bradford Teaching Hospitals NHS Foundation Trust and supported by the Assuring Autonomy International Programme at the University of York. The views expressed in this paper are those of the authors and not necessarily those of the NHS, or the Department of Health and Social Care. 

\bibliographystyle{ieeetr}
\bibliography{reference}

\end{document}